\documentclass[10pt,twocolumn,letterpaper]{article}

\usepackage{3dv}
\usepackage{times}
\usepackage{epsfig}
\usepackage{graphicx}
\usepackage{amsmath}
\usepackage{amssymb}
\usepackage{bbm}        
\usepackage{bm}
\usepackage{booktabs}
\usepackage{multirow}
\usepackage{pifont}
\usepackage{array}
\usepackage{enumitem}
\usepackage{pdfpages}

\usepackage[table,xcdraw,usenames, dvipsnames]{xcolor}
\usepackage{arydshln}	

\usepackage[pagebackref=true,breaklinks=true,colorlinks,bookmarks=false]{hyperref}

\newcolumntype{x}[1]{>{\centering\arraybackslash\hspace{0pt}}p{#1}}

\newcommand{\figref}[1]{Fig.~\ref{#1}}
\newcommand{\tabref}[1]{Table~\ref{#1}}

\newcommand{\cmark}{\textcolor{OliveGreen}{\ding{51}}}
\newcommand{\xmark}{\textcolor{red}{\ding{55}}}

\newcommand{\boldparagraph}[1]{\vspace{0.1em}\noindent{\bf #1} }

\DeclareMathOperator{\BCE}{BCE}

\threedvfinalcopy 

\def\threedvPaperID{381} 

\ifthreedvfinal\fi
\begin{document}

\title{RealisticHands: A Hybrid Model for 3D Hand Reconstruction}




\author{
%
Michael Seeber$^1$ \qquad
Roi Poranne$^2$ \qquad
Marc Polleyfeys$^{1,4}$ \qquad
Martin R. Oswald$^{1,3}$\\
$^1$ETH Zurich \quad 
$^2$University of Haifa \quad
$^3$University of Amsterdam \quad
$^4$Microsoft\\ 
\\
%
%
}

\maketitle
\thispagestyle{empty}

\begin{abstract}
    Estimating 3D hand meshes from RGB images robustly is a highly desirable task, made challenging due to the numerous degrees of freedom, and issues such as self-similarity and occlusions. Previous methods generally either use parametric 3D hand models or follow a model-free approach. While the former can be considered more robust, e.g. to occlusions, they are less expressive.
    We propose a hybrid approach, utilizing a deep neural network and differential rendering based optimization to demonstrably achieve the best of both worlds.
    In addition, we explore Virtual Reality (VR) as an application.
    Most VR headsets are nowadays equipped with multiple cameras, which we can leverage by extending our method to the egocentric stereo domain.
    This extension proves to be more resilient to the above mentioned issues.
    Finally, as a use-case, we show that the improved image-model alignment can be used to acquire the user's hand texture, which leads to a more realistic virtual hand representation.
\end{abstract}
\vspace{-3pt}
\section{Introduction}
Hand pose and shape estimation from images are long standing problems in computer vision.
Both are fundamental components in making mixed reality devices more immersive and accessible.
As mixed reality headsets are becoming ubiquitous, so are hands as a primary input device, replacing the old and clumsy controllers of the past.
However, in order to create the illusion of mixed reality, high degrees of performance and fidelity are necessary.

Our goal is to enable fast and accurate hand pose and shape estimation, including the hand texture too, in order to ultimately create a better sense of body ownership in virtual reality.
However, while both accuracy and processing speed are desirable in general, they are notoriously difficult to attain together.
In addition, small reconstruction errors are visibly amplified when textures are involved.

Previous work can be divided into model-based and model-free approaches.
While model-based methods are generally more robust to e.g. occlusions, they tend to be slower and sensitive to initialization.
Additionally, they are limited by the representation power of the underlying model, and cannot faithfully represent the whole diversity of hand shapes.
In contrast, model-free methods require vast amounts of data and many one-shot predictions often lead to poor image-model alignments.
Some methods rely on depth images, but with the emergence of deep learning, RGB images as a single input modality are becoming more popular, as such methods do not require special hardware.
Generally, this task is much more challenging, due to self-occlusions, finger self-similarity, and depth ambiguities.

\begin{figure}[tb]
  \vspace{-3pt}
  \centering
  \scriptsize
  \newcommand{\cellw}{66pt}
  \begin{tabular}{x{\cellw}x{\cellw}x{\cellw}}
     Input Image & Keypoints & Silhouette\\
  \end{tabular}
  \includegraphics[width=\linewidth]{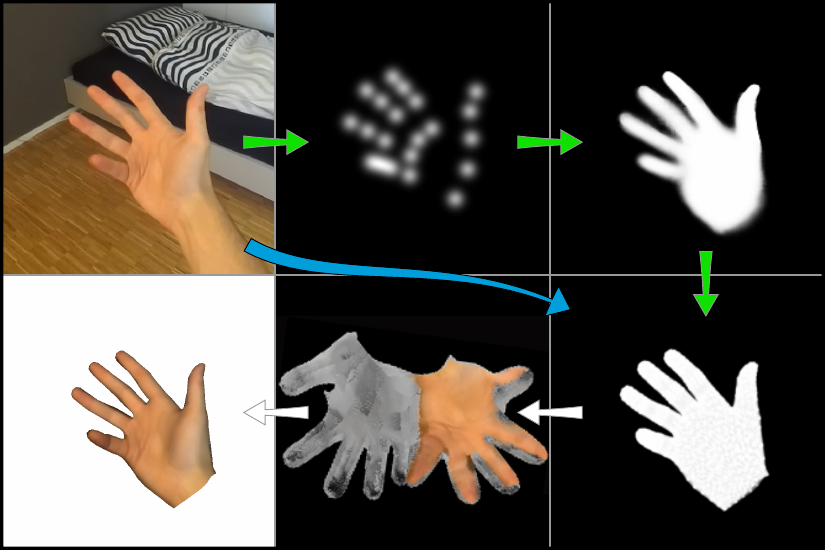} 
  \begin{tabular}{x{\cellw}x{\cellw}x{\cellw}}
    Textured 3D Mesh & Texture Map & 3D Mesh 
  \end{tabular}
  \caption{Given a single RGB image of a hand, our methods produces a faithful, textured 3D hand mesh. Our approach is hybrid between a \textcolor{Green}{model-based} approach (\textcolor{Green}{green path}) and a \textcolor{blue}{model-free} approach (\textcolor{blue}{blue path}).}
  \vspace{-12pt}
  \label{fig:teaser}
\end{figure}   

We propose a hybrid approach that combines the robustness of model-based methods with the expressiveness of model-free methods.
We use a deep learning approach to quickly and robustly obtain predictions, and a test-time optimization to further refine the result.
This enables easy texture extraction based on a straightforward projection.
In addition, we provide a stereo extension, making the method more robust, particularly for egocentric viewpoints.

To summarize, our \emph{contributions} are:
\textbf{(1)} Accurate 3D mesh estimation, fusing input modalities using a novel training strategy and improving resiliency by leveraging stereo views.
\textbf{(2)} Hand segmentation and fine grained fit optimization of the predicted MANO mesh via differential rendering.
\textbf{(3)} Personalized hand mesh rig including camera captured hand texture. 




\section{Related Work}
\boldparagraph{3D Hand Pose Estimation}
is the process of recovering the joints of a 3D hand skeleton, mainly from RGB and/or depth cameras.
Early methods \cite{athitsos2003estimating} used low-level visual cues such as edge maps and Chamfer matching~\cite{borgefors1988hierarchical} to rank the likelihood of possible predefined poses.
The advent of low-cost depth cameras paved the way for unconstrained 3D hand pose estimation and research focused on using depth data as input, with earlier methods based on optimization \cite{oikonomidis2011efficient, sharp2015accurate, tagliasacchi2015robust}.
Later, works such as \cite{zhang20163d,panteleris2017back} were able to replace depth by stereo cameras to a certain extent.

Convolutional Neural Networks (CNN) are a natural choice for pose estimation.
A popular approach is to predict keypoint joint locations.
Earlier methods, e.g. \cite{oberweger2015training}, used depth images as input and processed them with 2D CNNs.
Others argue that 2D CNNs are too limited, due to their lack of 3D spatial information~\cite{ge2016robust, ge20173d, moon2018v2v, ge2018hand}.
Normal hand motions exhibits a large range of viewpoints for a given articulation, which makes it difficult to define a canonical viewpoint.
To overcome this, various intermediate 3D representations have been studied (see e.g.~\cite{ge2016robust}).
Other methods employed D-TSDF~\cite{ge20173d}, voxels~\cite{moon2018v2v} or point sets~\cite{ge2018hand}.

More recent approaches can recover hand poses from RGB images only.
This is a significantly harder challenge due to the problem ill-posedness without depth information, but that has become feasible nonetheless thanks to learning based formulations e.g. \cite{zimmermann2017learning}.
One of the limiting factors however, is the massive amount of annotated data required in order to resolve the inevitable depth ambiguities.
Mueller~\etal~\cite{mueller2018ganerated} used generative methods to translate synthetic training data into realistic training images.
Due to the large discrepancy between factors of variation ranging from image background and pose to camera viewpoint, Yang~\etal~\cite{yang2019disentangling} introduced disentangled variational autoencoders to allow specific sampling and inference of these factors.
Another approach to deal with the lack of annotated training data is by utilizing a depth camera for weak supervision of the training procedure~\cite{cai2018weakly}.

\boldparagraph{3D Hand Reconstruction.}
While pose estimation yields a skeletal representation of the hand, hand reconstruction aims to recover the surface geometry of the hand.
For the more general full body reconstruction problem, multiple approaches have been proposed, such as 3D supervised learning methods (e.g. \cite{guler2018densepose}), or model-based deformable surface methods, like SMPL~\cite{loper2015smpl,kanazawa2018end,kolotouros2019learning}.
Inspired by the success of SMPL, a similar parameterized model for hands known as MANO was proposed in~\cite{MANO:SIGGRAPHASIA:2017}, and was followed up and used in many publications, such as~\cite{boukhayma20193d,zhang2019end,hasson19_obman,chen2021camera}.

Although MANO and such are able to represent a large variety of hand shapes, as a model-based approach, they are still limited in their expressiveness.
Model-free methods like \cite{ge20193d, choi2020pose2mesh} , use a graph CNNs to directly regress vertex positions in a coarse-to-fine manner. \cite{kulon2020weakly} relied on a spiral operator to construct spatial neighborhoods.
Another approach to obtain vertices directly was recently proposed  by Moon~\etal~\cite{moon2020i2l}, where they introduce a novel Image-to-Lixel (I2L) prediction network which retains the spatial relationship between pixels in the input image.
Additionally, to cope with the limited annotated data available also self-supervised methods \cite{chen2021model} have been studied.


\section{Method}

\boldparagraph{Overview.}
Our method estimates a textured 3D hand mesh from monocular or stereo inputs.
We go beyond simply estimating the hand model parameters only, by having an additional differential rendering-based optimization step for fine grained adjustments.
This makes our approach more expressive compared to solely relying on the MANO~\cite{MANO:SIGGRAPHASIA:2017} hand model and also leads to more accurate image-model alignments compared to other one-shot prediction methods.
This further enables us to rely on projections of the camera image to yield accompanying textures for a more personalized digital hand replication.
Furthermore we extend our method towards a stereo setting to cope better with occlusions an ambiguities and thus improve robustness.

\subsection{Architecture}

\begin{figure*}[t]
    \centering
    \includegraphics[width=\textwidth]{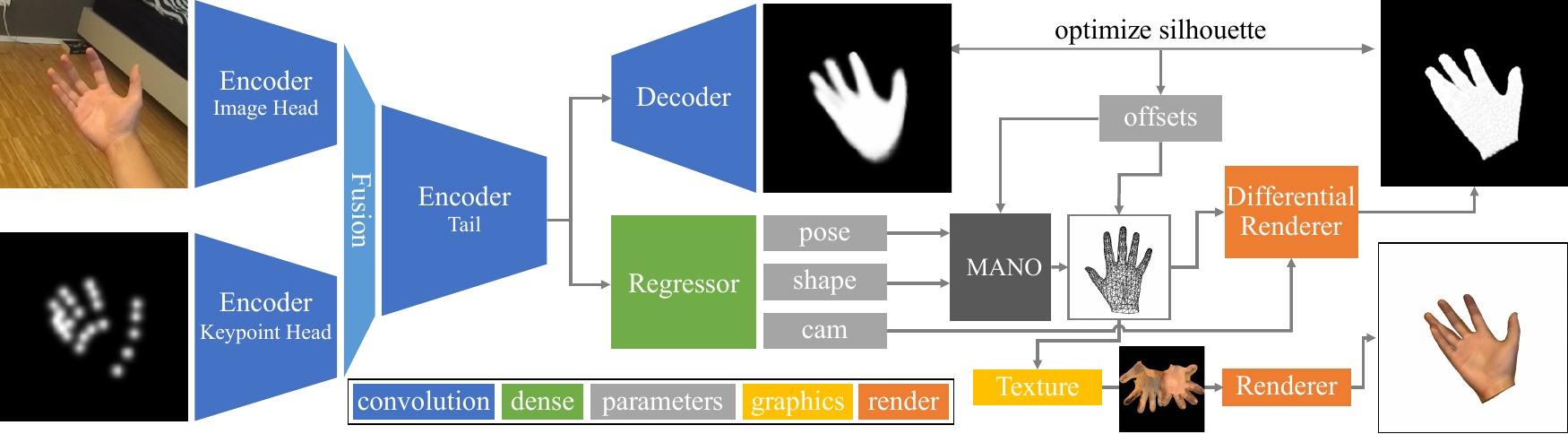}
    \caption[Overview of the architecture for a monocular camera setting]{\textbf{Overview of the architecture for a monocular camera setting.} }
    \label{fig:architecture_overview}
\end{figure*}

Our proposed network architecture for the monocular setting is illustrated in \figref{fig:architecture_overview}.
In short, inputs are fed into our deep encoder-decoder network to regress MANO and camera parameters and to predict a segmentation mask.
Then, the obtained hand mesh is optimized via differential rendering, such that the error between the segmentation mask and the rendered silhouette is minimized.

\boldparagraph{Input.}
The inputs consist of an RGB hand crop and hand keypoints encoded in a 21 channel heatmap, where each channel represents a joint.
We obtain them using MediaPipe Hands~\cite{zhang2020mediapipe}, a lightweight hand joint detection library.
The 2D joint locations are used to compute the bounding box for the hand crop \emph{and} the keypoint heatmap by encoding each joint location by a 2D Gaussian distribution.

\boldparagraph{Encoder.} 
We require an architecture with high representation capability, which is also computationally cheap enough for our targeted application.
Thus, we base our encoder on the ResNet50 architecture~\cite{he2016deep}.
Since we have an RGB image \emph{and} keypoint heatmaps, we extend the ResNet to support both modalities and included a learned implicit modality fusion.
We do this by dividing the network at the third ResNet layer into a front section and a tail section.
Then, we duplicate the head section for both inputs and adapt them to the corresponding input channel sizes.
To fuse the modalities in a sensible way, we add a fusion block that is inspired by the self-supervised model adaption block introduced in \cite{valada2019self}.
The idea behind the fusion block is to adaptively recalibrate and fuse encoded feature maps according to spatial location.
In other words, it learns to explicitly model the correlation between two feature maps, to selectively emphasize more informative features from a respective modality while suppressing the others.
This is achieved by concatenating the feature maps of each modality: image $\bm{X}^{img} \in \mathbb{R}^{C \times H \times W}$ and keypoints $\bm{X}^{kp} \in \mathbb{R}^{C \times H \times W}$ to $\bm{X}^{img|kp} \in \mathbb{R}^{2C \times H \times W}$.
Afterwards, $\bm{X}^{img|kp}$ is passed through a bottleneck, where the first ReLU activated convolution reduces the channel dimensionality by a ratio of $8$ to $2C/8 = 128$.
Subsequently, the dimensionality of the features is increased back to their original size by the second convolutional layer.
This time, a sigmoid activation is used to scale the values to the $[0,1]$ range, so that they represent weights.
The resulting output $w$ is used to emphasize and de-emphasize different regions in the originally concatenated features $\bm{X}^{img|kp}$ by taking the Hadamard product, i.e. performing element wise multiplication of the obtained weights and the concatenated features, which corresponds to $\hat{\bm{X}}^{img|kp} = w \circ \bm{X}^{img|kp} $.
As a last step, the adaptive recalibrated feature maps $\hat{\bm{X}}^{img|kp}$ are passed through a final convolution layer in order to reduce the channel depth and yield the fused output with the correct dimensionality, so that it can be further processed by the tail of the encoder to obtain the base features, of which the decoding components of our method make use.

\boldparagraph{Decoder.} 
 The decoder part of our architecture is composed out of two components, a MANO regressor and a segmentation decoder.
 The structure of the MANO regressor is straightforward and consists of dense layers only.
 The obtained base features $\bm{X}^{base} \in \mathbb{R}^{2048}$ from the encoder are flattened and processed by two fully connected layers with respective sizes $2048$ and $512$, from which the outputs are regressed.
 We predict the paramerters of the MANO hand model, which consist of pose $p \in \mathbb{R}^{48}$ and shape $s \in \mathbb{R}^{10}$.
 The pose parameters correspond to the global rotation and joint rotations in angle axis representation.
 The shape is determined through $10$ PCA components that encode the captured hand shape variance of the MANO hand scans \cite{MANO:SIGGRAPHASIA:2017}. By feeding $p$ and $s$ into the MANO model we obtain an articulated and shape adapted hand mesh $\in \mathbb{R}^{778 \times 3}$. 
 Lastly, we obtain weak-perspective camera parameters $c = (t, \delta) \in \mathbb{R}^3$, where $t \in \mathbb{R}^2 $ is the 2D translation on the image plane and $\delta \in \mathbb{R}$ is a scale factor.
 This allows to project a posed 3D hand model back onto the input image and obtain absolute camera coordinates.
 
 The segmentation decoder is tasked with attributing every pixel in the input hand crop to either 'hand' or 'background'.
 Like in \cite{long2015fully,ronneberger2015u}, we use an adapted U-Net with lateral connections between the contracting encoder and the successive expanding convolutional layers of the decoder.
 The first skip connections use the RGB branch of the encoder, as it contains more expressive segmentation features, but the segmentation decoder can benefit from the fused keypoint modality through deeper levels.

\subsection{Test-time Refinement}
Previous work show that one-shot MANO parameter regression methods~\cite{boukhayma20193d, zhang2019end, hasson19_obman}, although fast and robust, often have poor image-mesh alignment.
This is attributed to the pose being defined as relative joint rotations.
Therefore, minor rotation errors accumulated along the kinematic chain can result in noticeable drifts from 2D image features.
To mitigate this problem, we propose a test time optimization strategy.
The idea is to iteratively refine the hand mesh to better align with the input image during test-time, by adding offsets $\Delta p,\Delta s$ and $\Delta v$ for the pose and shape and vertices respectively.
Let $\bm{M}$ be the MANO model, the optimized mesh $\bm{H}$ at iteration step $t+1$ is then
\begin{equation}
\bm{H}_{t+1} = \bm{M}\left(p+ \Delta p_t, s + \Delta s_t\right) + \Delta v_t
\end{equation}
Similarly, we update $c$ by $c_{t+1} = c + \Delta c_t$.
All offsets $\Delta P_t = (\Delta{p}_t, \Delta{s}_t$, $\Delta{v}_t,\Delta{c}_t)$ are obtained from the gradients provided by a differential renderer, aiming to reduce the discrepancy between the predicted segmentation and the rendered silhouette of $\bm{H}_t$, i.e. make the rendered mesh silhouette $S_{mesh}$ more similar to the target silhouette image $S_{target}$ obtained from the segmentation decoder.

We use SoftRas~\cite{liu2019soft} as differential renderer, which aggregates probabilistic contributions of all mesh triangles with respect to the rendered pixels.
This formulation allows to propagate gradients to occluded and far-range vertices too.
We use stochastic gradient descent with the learning rate $\eta = 0.002$ and momentum $\alpha = 0.9$, and update the offsets by $\Delta P_{t+1} = \eta \nabla \mathcal{L}(H_t) + \alpha \Delta P_{t+1}$,
%
where $\mathcal{L}(H)$ is described in the following.

\boldparagraph{Refinement Loss.}
The loss $\mathcal{L}$ is a weighted sum of an \emph{image silhouette loss} and several regularization terms that are commonly used and thoroughly studied in literature.
First, the image silhouette loss is computed as the squared $L_2$ distance between the predicted silhouette and the target.
Namely $\mathcal{L}_\text{sil} = \left\| S_\text{mesh} - S_\text{target} \right\|_{2}^2$ where $S_\text{mesh}$ is the rendered silhouette and $S_\text{target}$ the predicted segmentation mask.
Additionally, we minimize the vertex offsets to jointly optimize the pose and the shape parameters.
As mentioned the other regularizers are commonly used and minimize the mesh normal variance and Laplacian to  encourage a smooth surface, as well as an edge loss to encourage uniform distribution of the mesh vertices. 
These are defined below, 
\begin{align}
    \mathcal{L}_\text{v} &= \left\|\Delta v\right\|_{2}^{2},\quad \mathcal{L}_{\text{e}} = \sum_{v} \sum_{k \in \mathcal{N}(v)}\|v-k\|_{2}^{2} \\
    \mathcal{L}_\text{lap} &= \sum_{v}\left\|\delta_{v}\right\|_{2}^{2} ~ \text{, where }
    \delta_{v}=v\!-\!\!\!\!\!\sum_{k \in \mathcal{N}(v)} \!\!\frac{k}{\|\mathcal{N}(v)\|} \\
    \mathcal{L}_{n} &= 1 - \frac{n_l \cdot n_r}{||n_l|| \cdot ||n_r||}   
\end{align}
where $v$ corresponds to a vertex, $\mathcal{N}(v)$ are the edges connected to $v$, and $n_l,n_r$ are normals of neighboring faces. Finally $\mathcal{L}$ is formally defined by
\begin{equation}
\label{eq:method_offset_loss}
 \mathcal{L} = 
\underbrace{
 	\lambda_1 \mathcal{L}_\text{sil}
}_\text{image loss} \!+\!
\underbrace{
    \lambda_2 \mathcal{L}_\text{v} \!+\! 
 	\lambda_3 \mathcal{L}_\text{n} \!+\!
	\lambda_4 \mathcal{L}_\text{lap} \!+\!
	\lambda_5  {\mathcal{L}}_\text{edge}
}_\text{regularize /smooth loss}
\end{equation}
where we used $\lambda_1, \lambda_2, \lambda_3, \lambda_4 = 1$ and $\lambda_5 = 0.1$.

\boldparagraph{Texturing.}
To obtain a texture, we unwrap the MANO hand mesh to find UV coordinates.
During test time, we rasterize the UVs over the mesh to obtain UV coordinates per fragment, which can be used to map colors from the image to the texture map.
This is done in every frame, to adapt to changes in the texture caused by e.g. wrinkling of the skin.
To fight outliers we use exponential smoothing when updating the texture, i.e. we average the current texture with the previous.


\begin{figure}[tb]
 \centering
 \includegraphics[width=\linewidth]{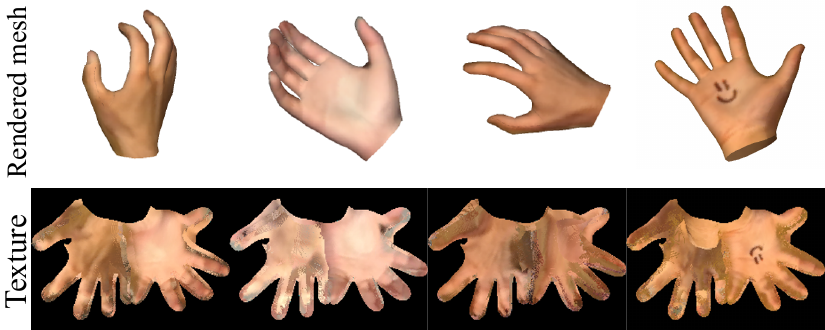}
 \caption{\textbf{Visualization of texture maps and rendered meshes.}}
 \label{fig:meshplustexture}
 \vspace{-10pt}
\end{figure}

\subsection{Stereo Extension}
\label{sec:method_stereo}
Estimating a 3D hand pose from a single RGB image is an ill-posed problem due to depth and scale ambiguities.
Stereo images have the ability to resolve these, but stereo training data is lacking.
Thus we propose a network architecture that allows the use of monocular datasets like FreiHAND to learn the correlations between poses and real hand image data.
Limited stereo data such as our proposed synthetic hand dataset, introduced in Sec.~\ref{sec:method_datasets} can then be utilized to learn a sensible stereo fusion.


We extend our model to support stereo images, mostly by duplicating the mono architecture. An illustration of the full architecture can be found in the supplementary material.
However, in addition to regressing $p,s$ and $c$ for each view from the base features $\bm{X} \in \mathbb{R}^{512} $ of the MANO regressor, we also concatenate them from both views into $\bm{X}_{stereo} \in \mathbb{R}^{1024} = \bm{X}_{right} | \bm{X}_{left}$.
From this, we regress additional stereo weights $w \in \mathbb{R}^{48}$ using a fully connected layer.
The obtained weights are used to combine the predicted poses from the left and right views by computing the new stereo fused right hand pose $p_\text{right}$ as
\begin{equation}
   p_\text{right} = w ~ p_\text{right} +  (1-w) ~  p_\text{left}.
\end{equation}
In other words, we ideally want the network to be able to differentiate visible and occluded joints of different views, so that they can be merged in a meaningful way.
For example, if a joint is self-occluded by the palm in the right view, the exact location is not recoverable and we have to rely on predicting likely distributions of possible locations.
However, if the specific joint is visible in the left view, the network now has the power to compensate for the lack of information by utilizing the left view joint predictions over the right view predictions. 

The predicted shape parameters from both views are fused by averaging, the same way we refine the left and right camera parameters while taking their geometric constraints into account.We opted to not learn weights for fusing these parameters, because of the small variance exhibited between the views.
With $\xi$ transforming from left to right view, we obtain
\begin{equation}
   s_\text{stereo} = \frac{s_\text{right} + s_\text{left}}{2} \qquad
   c_\text{right} = \frac{c_\text{right} + \xi(c_\text{left})}{2}.
\end{equation}
We can also leverage stereo during the optimization stage, by computing the silhouette image loss as the average over the two views. 
Letting $S^L,S^R$ be the left and right silhouette images, the stereo loss term is simply
\begin{figure*}[t!]
    \vspace{-8pt}
    \centering
    \includegraphics[width=\textwidth]{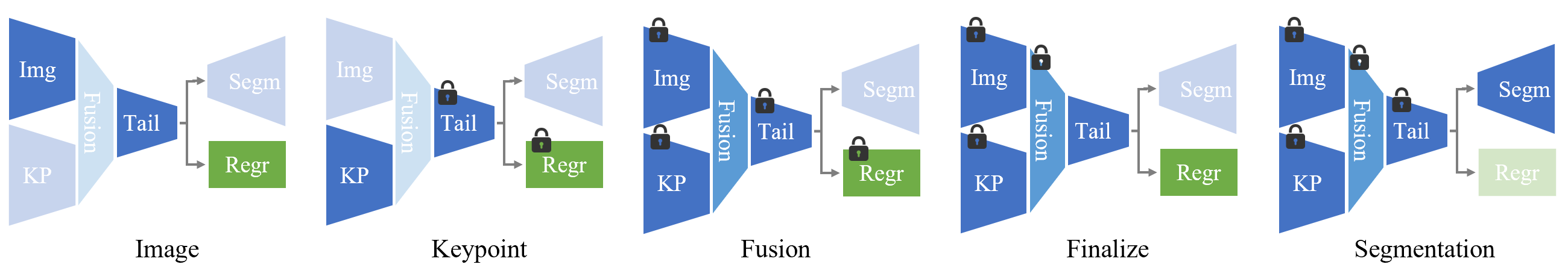}\\[-4pt]
    \caption[Trainable components]{\textbf{Training framework.} Visualization of the five monocular training stages. The active components of each stage are highlighted. Components with the lock symbol have their parameters fixed, i.e. they are not updated during training. }
    \label{fig:method_training}
\end{figure*}
\begin{equation}
    \mathcal{L}_\text{sil} = \frac{\left\| S^L_\text{mesh} - S^L_\text{target} \right\|_{2}^2 + \left\| S^R_\text{mesh} - S^R_\text{target} \right\|_{2}^2}{2}
\end{equation}

\subsection{Datasets}
\label{sec:method_datasets}

\boldparagraph{FreiHAND Dataset.}
The FreiHAND~\cite{zimmermann2019freihand} dataset consists of 32,560 unique RGB images of right hands with the corresponding MANO annotation, intrinsic camera parameters and 2D joint locations. The FreiHAND dataset is unique in that the annotations include not only pose but also shape parameters for the MANO model.


\boldparagraph{Synthetic Stereo Hands Dataset.}
Since training data for a stereo setting with focus on egocentric viewpoints is currently lacking, we generated a large-scale synthetic dataset.
The primary use is to learn the proposed stereo fusion, but the pixel-accurate segmentation masks are also utilized for training the segmentation decoder.
We adapt ObMan \cite{hasson19_obman} to our needs by modeling our stereo camera and focusing on egocentric viewpoints. We render left and right RGB images, corresponding segmentation masks and output annotations that include hand vertices, 2D joints, 3D joints along with the camera parameters.
In total, the dataset contains $15082$ samples, corresponding to $30164$ RGB images, annotations and segmentation masks.

\subsection{Training Framework}
\label{sec:method_training_framework}
Training is performed in multiple stages, where different components of the model are frozen or removed components from the information flow.
This has proven to greatly increase performance, as discussed in 
\tabref{tab:results_training_stages}, at a marginal implementation overhead.
For the monocular model we used five training stages, as illustrated in Fig. \ref{fig:method_training}.
The `Image', `Keypoint', `Fusion' and `Finalize' stage use the FreiHAND dataset with data augmentation (Sec. \ref{sec:method_traininglosses}).
The `Segmentation' stage use both our synthetic stereo hands dataset, which is pixel-perfect but synthetic, and FreiHAND, which is real but less accurate).
This is done in order to mitigate the drawback of each respective dataset.

\begin{figure}[tb]
 \vspace{-8pt}
 \centering
 \includegraphics[width=\linewidth]{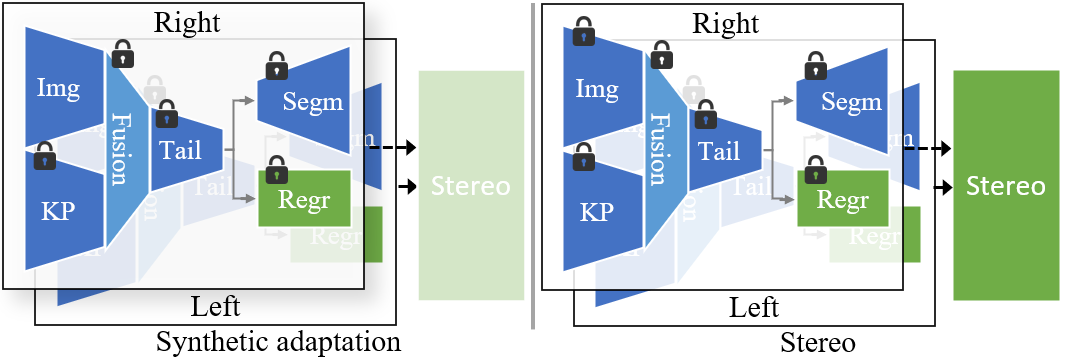}
 \caption{\textbf{Visualization of the two stereo training stages.} In the synthetic adaption stage the image head is retrained, whereas in the stereo stage the fusion weights are learned.}
 \label{fig:method_training_stereo}
\end{figure}

The stereo setting includes two additional stages, which are carried out using the synthetic stereo hand dataset (Fig.~\ref{fig:method_training_stereo}).
The first `synthetic adaption' stage retrains the image encoder to adapt towards the synthetic image data.
We conclude by training the `Stereo' weights regressor to learn a sensible fusion.
During test time on real data, the original non synthetically adapted image encoder is used.

Additionally, the training strategy does not only outperform a joint training that tends to emphasize the dominant branch, but also exhibits better adaptability and flexibility, as only the image head could be retrained to adapt to a new setting.

\subsection{Training Losses}
\label{sec:method_traininglosses}

\boldparagraph{MANO Loss.}
We use the following weighted loss to train the network that regresses the MANO parameters:
\begin{equation}
     \mathcal{L}_\text{MANO} = 
 	\underbrace{
		\lambda_1 \mathcal{L}_\text{pose} \!+\!
		\lambda_2 \mathcal{L}_\text{shape} \!+\!
		\lambda_3 \mathcal{L}_\text{joints}^\text{2D}
	}_\text{mano losses} \!+\!
	\underbrace{
		\lambda_4 \mathcal{L}_\text{shape} \!+\!
		\lambda_5 \mathcal{L}_\text{cam}
	}_{\text{regularization}}
	\label{eq:mano_loss}
\end{equation}
where  $\mathcal{L}_\text{pose}, \mathcal{L}_\text{shape},\mathcal{L}_\text{joints}^{2D},\mathcal{L}_\text{cam}$ are the MSE of the pose, shape, 2D joints and scale camera parameters,
 and $\mathcal{L}_\text{shape}$ is the norm of the shape PCA.
%
While $\mathcal{L}_\text{pose}$ and $ \mathcal{L}_\text{shape}$ infer MANO parameters, $\mathcal{L}_\text{joints}^{2D}$ learns the weak camera parameters for the root joint, so that the mesh can be projected back onto the image.
As the MANO model encodes the mean shape by the zero vector, we introduce an additional regularization term $\mathcal{L}_\text{shape}$ to prevent blow ups.
Finally, we add $\mathcal{L}_\text{cam}$ to avoid placing the root joint behind the camera.
We used  $\lambda_1 = 10$, $\lambda_2 = 1$  $\lambda_3 = 100$, $\lambda_4 = 0.5$ and $\lambda_5 = 10^{-5}$.

\boldparagraph{Segmentation Loss.}
For training the segmentation decoder, we rely on the binary cross entropy.
\begin{equation}
     \mathcal{L}_\text{seg} = \BCE(\hat{y}, y) = -(y \log(\hat y) + (1-y) \log(1- \hat{y}))
\end{equation}
\boldparagraph{Stereo Loss.}
We use a weighted sum similar to Eq.~\eqref{eq:mano_loss},
but we directly supervise the vertices and the 3D joint locations by minimizing their corresponding MSE's $\mathcal{L}_\text{vert}$ and $\mathcal{L}_\text{joints}^{3D}$:
\begin{equation}
     \mathcal{L}_\text{stereo} = 
 	\lambda_1 \mathcal{L}_\text{vert} +
 	\lambda_2 \mathcal{L}_\text{joints}^{3D}
\end{equation}
%

\boldparagraph{Data Augmentation.}
We use common techniques such as rescaling, rotating, cropping and blurring during training.


%
\begin{table}[b]
  \vspace{-8pt}
  \centering
  \footnotesize
  \setlength{\tabcolsep}{2.4pt}
\begin{tabular}{@{}llllll@{}}
\toprule
                                                 & MPVE             & F@5mm          & F@5mm          & Model   & GT     \\
Methods                                          & (PA)$\downarrow$ & (PA)$\uparrow$ & (PA)$\uparrow$ & based   & scale  \\ \midrule
Mean shape                                       & 1.64          & 0.336          & 0.837          & \cmark  &        \\
Inverse Kinematics~\cite{zimmermann2019freihand} & 1.37          & 0.439          & 0.892          & \cmark  &        \\
Hasson~\etal~\cite{hasson19_obman}               & 1.33          & 0.429          & 0.907          & \cmark  & \cmark \\
Boukhayma~\etal~\cite{boukhayma20193d}           & 1.32          & 0.427          & 0.894          & \cmark  & \cmark \\
Mano CNN~\cite{zimmermann2019freihand}           & 1.09          & 0.516          & 0.934          & \cmark  & \cmark \\
Kulon~\etal~\cite{kulon2020weakly}           & 0.86         & 0.614          & 0.966          & \xmark  & \xmark \\
I2L~\cite{moon2020i2l}                           & 0.76          & 0.681          & 0.973          & \xmark  & \xmark \\  \hline
\textbf{Ours} (Mono, MP Hands)                   & 0.97 & 0.575 & 0.949 & \cmark  & \xmark     \\
\textbf{Ours} (Mono, PoseNet(I2L))               & 0.78 & 0.662 & 0.971 & \cmark  & \xmark     \\ \bottomrule
\end{tabular}
\vspace{2pt}
  \caption{\textbf{Quantitative results on the FreiHAND benchmark test set.} Comparison of our approach with other methods on the task of monocular hand pose and shape estimation. Our method outperforms the other model based methods in all three evaluation metrics and shows similar, but qualitatively more robust performance than model free approaches. }
  \label{tab:results_test_monocular}
\end{table}

\section{Results}

All our experiments were carried out on a workstation with an AMD Ryzen 9 5900x, 64 GB of RAM and an Nvidia GTX TITAN X.
Our approach was implemented in PyTorch and the code, models and also our synthetic stereo dataset will be released upon publication. 

\boldparagraph{Metrics.}
The hand vertex prediction is evaluated by the Mean Per Vertex Position error (MPVE), which measures the Euclidean distance in millimeters between the estimated and groundtruth vertex coordinates.
We also report the Area Under Curve (AUC)~\cite{zimmermann2019freihand} of the Percentage of Correct Keypoints (PCK) by computing the percentage of predicted vertices lying within a spherical threshold around the target vertex position. 
Further, we report F-scores which - given a distance threshold - define the harmonic mean between recall and precision between two point sets~\cite{knapitsch2017tanks}.
To evaluate the segmentation performance we use pixel accuracy and the more meaningful mean intersection over union metric.
Also binary masks, obtained by the projection of the hand meshes are evaluated using the mean IoU.

\subsection{MANO Regression}
\label{sec:results_manoregression}

We evaluate the HandNet component which regresses MANO parameters from monocular RGB images.
The HandNet was trained for a total of 500 epochs, distributed with 120 + 100 + 100 + 180 epochs on the respective training stages (Sec. \ref{sec:method_training_framework}).
During the last 20 epochs of each stage the learning rate was linearly decayed. The Adam optimizer~\cite{kingma2014adam} was used with an initial learning rate of $0.0002$ at each stage and $\beta_1 =0.9$ and $\beta_2 = 0.99$.

\begin{figure}[t]
    \centering
    \scriptsize                         
    \setlength{\tabcolsep}{1pt}         
	\renewcommand{\arraystretch}{0.8}   
	\newcommand{\sz}{0.24}              
	\begin{tabular}{lcccc}
	    \rotatebox{90}{\hspace{20pt} Ours} &
        \includegraphics[width=\sz\columnwidth]{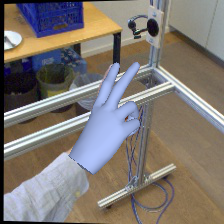} &
        \includegraphics[width=\sz\columnwidth]{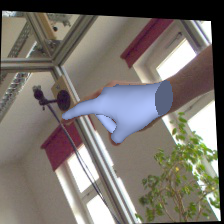} &
        \includegraphics[width=\sz\columnwidth]{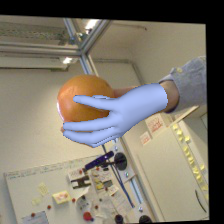} &
        \includegraphics[width=\sz\columnwidth]{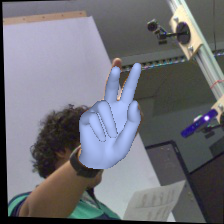} \\
        \rotatebox{90}{\hspace{18pt} I2L~\cite{moon2020i2l}} &
        \includegraphics[width=\sz\columnwidth]{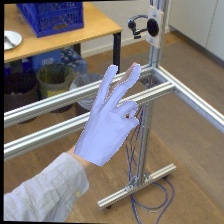} &
        \includegraphics[width=\sz\columnwidth]{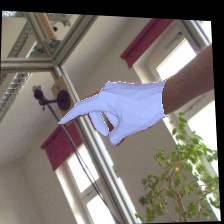} &
        \includegraphics[width=\sz\columnwidth]{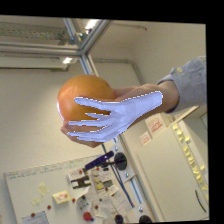} &
        \includegraphics[width=\sz\columnwidth]{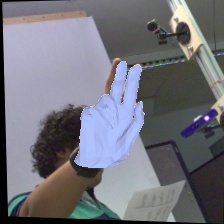}\\
    \end{tabular}    
    \caption{\textbf{Qualitative results on the FreiHAND test set.} Visualization of predictions on the FreiHAND test set, in comparision to I2L~\cite{moon2020i2l}. We used 3 optimization iterations for our method.}
    \label{fig:results_qualitative_freihand}
\end{figure}

\begin{figure}[tb]
 \centering
 \includegraphics[width=\linewidth]{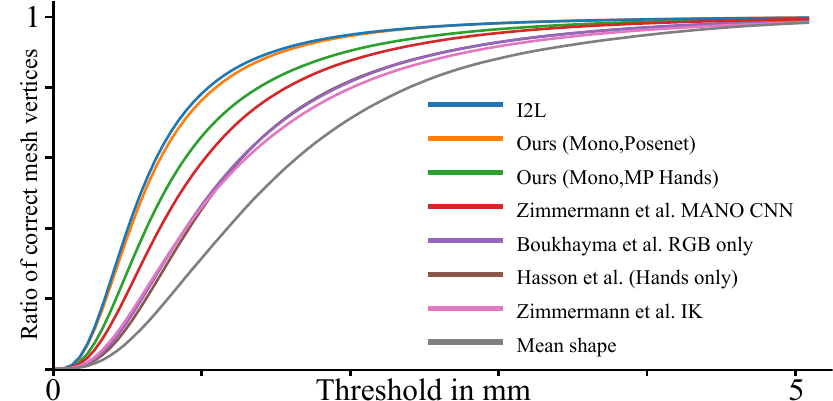}
 \caption{\textbf{PCK curves for the FreiHAND benchmark test.} Based on vertex positions using PA. We outperform all other model based approaches and show similar quantitative performance than the state of the art model free approach, while being more robust.}
 \label{fig:results_test_mesh}
\end{figure}

\boldparagraph{State-of-the-art Comparison.}
We compared the monocular HandNet component to the state-of-the-art using the FreiHAND test set, which provides no groundtruth, via an online challenge\footnote{\url{https://competitions.codalab.org/competitions/21238}}. 
As some methods predict only relative coordinates, the benchmark reports through Procrustes analysis (PA) aligned metrics.
Table~\ref{tab:results_test_monocular} shows that our approach outperforms baselines, such as the mean shape and inverse kinematic fits of the MANO model, and also ranks higher than other model based approaches such as \cite{hasson19_obman, boukhayma20193d, zimmermann2019freihand, kulon2020weakly}.
Further, it performs quantitative similarly to I2L \cite{moon2020i2l}, a model-free approach that occassionaly produces collapsed results (see Fig.~\ref{fig:results_qualitative_freihand}). The qualitative superiority of our methods over I2L is also well demonstrated in the supplementary video.
Note that I2L uses a 3D joint prediction network (PoseNet) trained on the FreiHand dataset, which is geared towards the test data more than our general MP Hands keypoint detector~\cite{zhang2020mediapipe} .
For a fair comparison, we use projected PoseNet 3D joint predictions as our keypoint input.
The PCK curves of the methods are visible in Fig.~\ref{fig:results_test_mesh}.

\boldparagraph{Modality Fusion.}
For the MANO Regression experiments we use a local validation set that was created by randomly splitting 30\% of the FreiHand samples into a validation dataset using the remaining 70\% for training the network. We made sure to base the training and test splits on unique samples of the FreiHAND dataset and include the different color-hallucinated versions in the respective set. 

\begin{figure}[tb]
\vspace{-5pt}
 \centering
 \includegraphics[width=\linewidth]{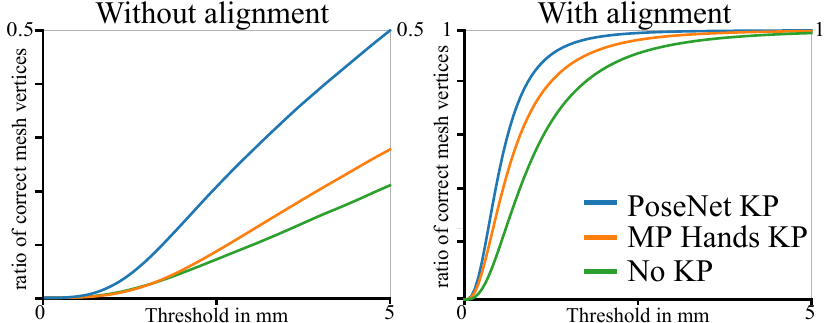}
 \caption{\textbf{Modality Fusion.} PCK plots comparing the performance of our method, when no keypoints are provided against using MP Hands \cite{zhang2020mediapipe} or PoseNet \cite{moon2020i2l} as keypoint detector on the validation dataset. }
 \label{fig:results_mesh_withKPwithout}
\end{figure}

To test the robustness of our method, when no keypoints are available, we evaluated the impact of the modality as well as the performance of the fusion component, in such cases. Therefore no test-time refinement was performed for this experiment. The results in \figref{fig:results_mesh_withKPwithout} with mesh alignment through PA, show that good keypoint prediction boost the performance and are a useful modality. When no meaningful information is provided by this input branch our fusion component successfully learned to suppress the keypoint modality with slightly lower performance.
In contrast, when we inspect the non-aligned results in \figref{fig:results_mesh_withKPwithout}, it becomes apparent that the keypoint heatmap provides especially valuable information for the global image-model alignment with larger difference between the versions.
This proves, that the additional modality and fusion are indeed helpful, because good initial global alignments are crucial for the fine grained optimization to succeed.

\boldparagraph{Training Framework.}
To evaluate the usefulness of the multi-stage training procedure we compare the performance to a version of the same network without the introduced training stages on the validation set. The results are shown in \tabref{tab:results_training_stages}.
The network trained using our training stages significantly outperforms a directly trained network and the ablation on the keypoint modality shows, that the fusion component is better capable at adapting to the provided inputs when trained with the proposed stages.

\begin{table}[tb]
\centering
\footnotesize
\setlength{\tabcolsep}{3.4pt}
\newcommand{\gcs}{\phantom{a}}  
\begin{tabular}{@{}lllcllcll@{}}
\toprule
    & \multicolumn{2}{c}{MPVE$\downarrow$}  & \gcs & \multicolumn{2}{c}{F@5mm$\uparrow$}  & \gcs & \multicolumn{2}{c}{F@15mm$\uparrow$}  \\[-1pt]
    \cmidrule{2-3} \cmidrule{5-6} \cmidrule{8-9} \noalign{\vskip -1pt}
            & no KP         & with KP       && no KP         & with KP       && no KP         & with KP       \\
            \midrule
single-stage   & 1.79          & 1.16          && 0.31          & 0.49          && 0.82          & 0.92          \\
multi-stage & \textbf{0.85} & \textbf{0.78} && \textbf{0.45} & \textbf{0.66} && \textbf{0.96} & \textbf{0.97} \\ \bottomrule
\end{tabular}
\vspace{1pt}
\caption{\textbf{Training stages.} The multi-stage training yields the best results when the keypoints are provided, but also outperforms the directly trained version with no keypoints. Especially the adaptive fusion seems to profit from training framework, as the network better learns to emphasize the respective modality.}
\label{tab:results_training_stages}
\end{table}

\subsection{Optimization}

\begin{figure}[tb]
\vspace{-8pt}
\centering
\footnotesize
\begin{minipage}{0.44\columnwidth}
\setlength{\tabcolsep}{1pt}
\renewcommand{\arraystretch}{1.0}
\begin{tabular}{@{}lll@{}}
\toprule
    & \multicolumn{2}{c}{IoU$\uparrow$} \\[-1pt] \cmidrule(l){2-3} \noalign{\vskip -1pt}
    & {\scriptsize MP Hands} & {\scriptsize PoseNet}     \\ \midrule
no refinement          & 0.677               & 0.722       \\
3 iterations           & 0.706               & 0.758       \\
10 iterations          & 0.735               & 0.793       \\
15 iterations          & \textbf{0.748}      & \textbf{0.806}  \\[0.3pt]
\hdashline \noalign{\vskip 2pt}
%
I2L~\cite{moon2020i2l} &                     & 0.737       \\ \bottomrule
\end{tabular}
\end{minipage}%
\begin{minipage}{0.56\columnwidth}
    \centering
    \scriptsize                         
    \setlength{\tabcolsep}{1pt}         
	\renewcommand{\arraystretch}{0.8}   
	\newcommand{\sz}{0.3}              
	\begin{tabular}{lccc}
        \rotatebox{90}{\hspace{8pt} Ours} &
        \includegraphics[width=\sz\columnwidth]{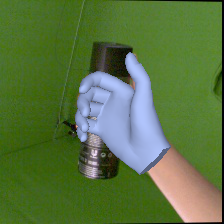} &
        \includegraphics[width=\sz\columnwidth]{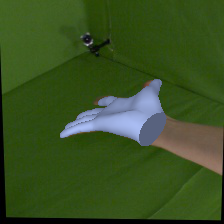} &
        \includegraphics[width=\sz\columnwidth]{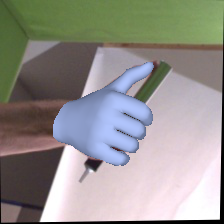} \\
        \rotatebox{90}{\hspace{4pt} I2L~\cite{moon2020i2l}} &
        \includegraphics[width=\sz\columnwidth]{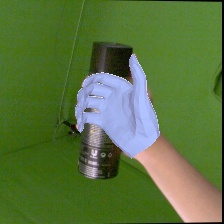} &
        \includegraphics[width=\sz\columnwidth]{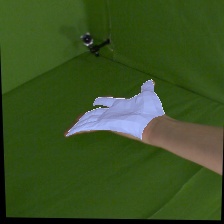} &
        \includegraphics[width=\sz\columnwidth]{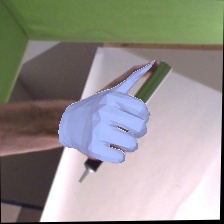} \\
    \end{tabular}    
\end{minipage}
\vspace{1pt}
\caption{\textbf{Projection error.} The table lists the IoU scores for the projection error of the hand mesh with the GT mask, when no optimization, 3, 10 and 15 optimization iterations are used. We further compare against I2L \cite{moon2020i2l}, where use the same PoseNet keypoint predictions. Additionally we visualize samples, where the projection error between our method (3 iterations) and I2L is comparable, but our method outputs qualitatively more robust hand meshes, compared to the partly collapsed I2L outputs.}
\label{fig:results_projection_results}
\end{figure}

To evaluate the benefit of the optimization stage, which refines the image-model alignment, we project the obtained mesh onto the image plane and compare it with the groundtruth segmentation mask. 
The validation results are reported in Fig.~\ref{fig:results_projection_results}, where we also compare against I2L \cite{moon2020i2l}. Additionally, we visualize samples with similar projection losses for both methods, but with qualitative more robust results produced by our approach.

\begin{figure}[tb]
    \vspace{-6pt}
    \centering
    \scriptsize                         
    \setlength{\tabcolsep}{1pt}         
	\renewcommand{\arraystretch}{0.8}   
	\newcommand{\sz}{0.23}              
	\begin{tabular}{lcccc}
	    \rotatebox{90}{\hspace{20pt} Input} &
        \includegraphics[width=\sz\columnwidth]{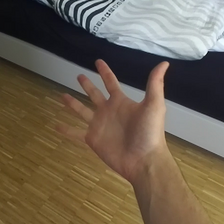} &
        \includegraphics[width=\sz\columnwidth]{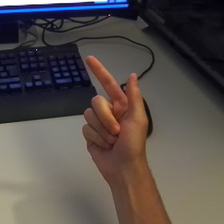} &
        \includegraphics[width=\sz\columnwidth]{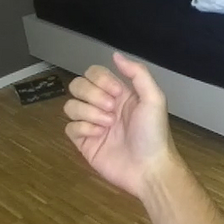} &
        \includegraphics[width=\sz\columnwidth]{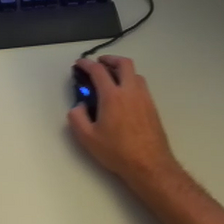} \\
        \rotatebox{90}{\hspace{20pt} Ours} &
        \includegraphics[width=\sz\columnwidth]{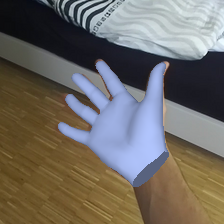} &
        \includegraphics[width=\sz\columnwidth]{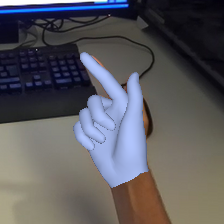} &
        \includegraphics[width=\sz\columnwidth]{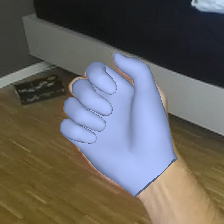} &
        \includegraphics[width=\sz\columnwidth]{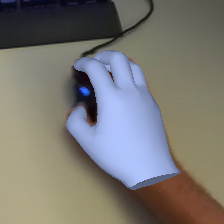} \\
        \rotatebox{90}{\hspace{15pt} I2L~\cite{moon2020i2l}} &
        \includegraphics[width=\sz\columnwidth]{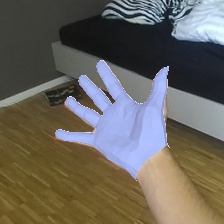} &
        \includegraphics[width=\sz\columnwidth]{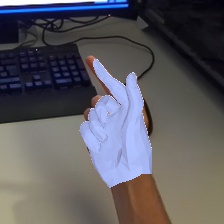}&
        \includegraphics[width=\sz\columnwidth]{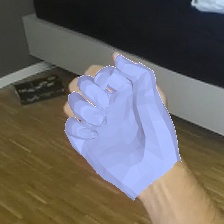} &
        \includegraphics[width=\sz\columnwidth]{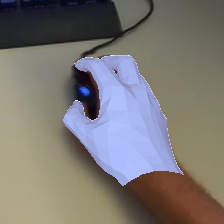} \\
    \end{tabular}    
    \caption{\textbf{Qualitative results on real captured data.} Visualization of predictions from real captured data, different from the FreiHand training data for both our method and I2L~\cite{moon2020i2l} as comparison. We used 3 optimization iterations for our method. The required root joint input for I2L is extracted from our predictions.}
    \label{fig:results_qualitative}
\end{figure}

\boldparagraph{Qualitative results.}
To compare our method qualitative with I2L~\cite{moon2020i2l} on real data different from the FreiHAND~\cite{zimmermann2019freihand} dataset, we captured video sequences with various hand poses.
Sample hand crops and the respective hand mesh predictions are visualized in Fig.~\ref{fig:results_qualitative}.
Our method is more robust, especially with fingers located on top of the palm.

\subsection{Segmentation}
The segmentation decoder was trained for 10 epochs, linearly decaying the learning rate during the last 6 epochs.
The ADAM optimizer \cite{kingma2014adam} was used with an initial learning rate of $0.0002$ and $\beta_1 =0.9$ and $\beta_2 = 0.99$.

The segmentation experiments used both FreiHAND and our synthetic dataset.
Specifically, we use 70\% of the data from each dataset for training and 30\% for validation.
Due to the chosen size of the synthetic dataset, the corresponding number of samples from the two datasets are also balanced.
\begin{figure}[t]
    \vspace{-6pt}
    \centering
    \scriptsize                         
    \setlength{\tabcolsep}{1pt}         
	\renewcommand{\arraystretch}{0.8}   
	\newcommand{\sz}{0.192}             
	\begin{tabular}{ccccc}
        \includegraphics[width=\sz\columnwidth]{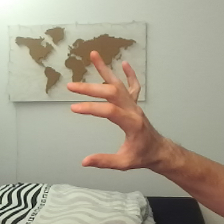} &
        \includegraphics[width=\sz\columnwidth]{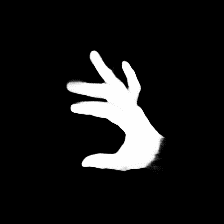} &
        \includegraphics[width=\sz\columnwidth]{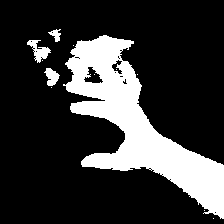} &
        \includegraphics[width=\sz\columnwidth]{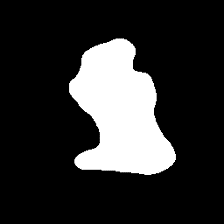} &
        \includegraphics[width=\sz\columnwidth]{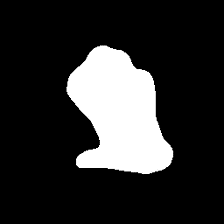} \\
        \includegraphics[width=\sz\columnwidth]{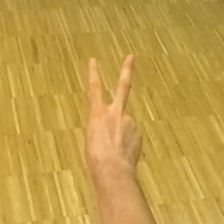} &
        \includegraphics[width=\sz\columnwidth]{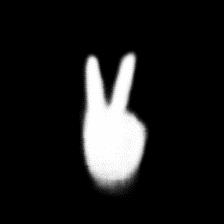} &
        \includegraphics[width=\sz\columnwidth]{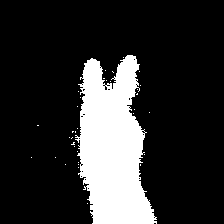} &
        \includegraphics[width=\sz\columnwidth]{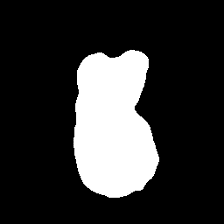} &
        \includegraphics[width=\sz\columnwidth]{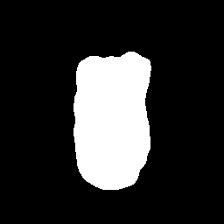} \\
        \includegraphics[width=\sz\columnwidth]{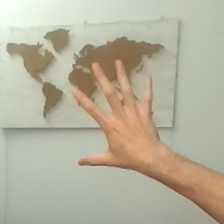} &
        \includegraphics[width=\sz\columnwidth]{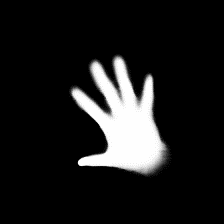} &
        \includegraphics[width=\sz\columnwidth]{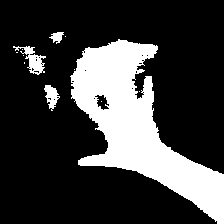} &
        \includegraphics[width=\sz\columnwidth]{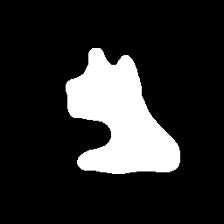} &
        \includegraphics[width=\sz\columnwidth]{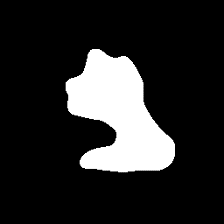} \\
        \includegraphics[width=\sz\columnwidth]{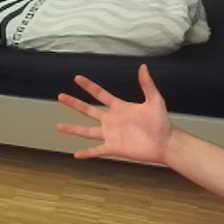} &
        \includegraphics[width=\sz\columnwidth]{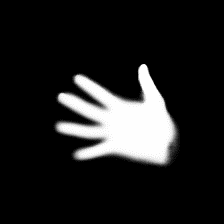} &
        \includegraphics[width=\sz\columnwidth]{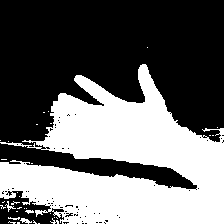} &
        \includegraphics[width=\sz\columnwidth]{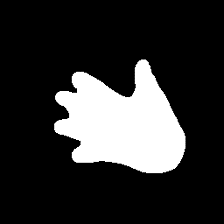} &
        \includegraphics[width=\sz\columnwidth]{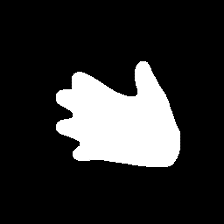} \\
        Input & Ours & HSV & CNN~\cite{cai2020generalizing} & UMA~\cite{cai2020generalizing}\\[2pt]
    \end{tabular}    
    \caption[Qualitative segmentation comparison]{\textbf{Qualitative segmentation comparison.} Visualization of obtained segmentations by different methods. HSV corresponds to thresholding on skin color in the HSV color space. CNN~\cite{cai2020generalizing} corresponds to the Bayesian CNN that serves as starting point for the uncertainty-guided model adaptation (UMA~\cite{cai2020generalizing}). }
    \label{fig:segmentation}
\end{figure}
\begin{table}[t]
\vspace{-6pt}
\centering
\footnotesize
\setlength{\tabcolsep}{2.9pt}
\renewcommand{\arraystretch}{1.0}
\newcommand{\gcs}{\phantom{a}}  
\begin{tabular}{@{}lcccccccc@{}}
\toprule
\multicolumn{1}{c}{\multirow{2}{*}{}} & \multicolumn{2}{c}{Combined} & \gcs & \multicolumn{2}{c}{FreiHAND} & \gcs & \multicolumn{2}{c}{Synth}   \\[-1pt]
\cmidrule{2-3}  \cmidrule{5-6} \cmidrule{8-9} \noalign{\vskip -1pt}
\multicolumn{1}{c}{}           & IoU$\uparrow$ & Accuracy$\uparrow$ && IoU$\uparrow$ & Accuracy$\uparrow$ && IoU$\uparrow$ & Accuracy$\uparrow$ \\ 
\midrule
HSV                            & 0.26     & 0.81     && 0.33     & 0.90     && 0.18     & 0.72     \\
CNN~\cite{cai2020generalizing} & 0.85     & 0.98     && 0.77     & 0.97     && 0.93     & 0.99     \\
UMA~\cite{cai2020generalizing} & 0.86     & 0.98     && 0.78     & 0.98     && 0.93     & 0.99     \\
Ours                           & \textbf{0.88} & \textbf{0.99} && \textbf{0.80} & \textbf{0.98} && \textbf{0.95} & \textbf{0.99} \\ \bottomrule
\end{tabular}
\vspace{1pt}
\caption{\textbf{Segmentation Results.} Our method performs best.}
\label{tab:results_seg}
\vspace{-5pt}
\end{table}

We compared the segmentation performance to the following three baselines, abbreviated by HSV, CNN, UMA.
\boldparagraph{HSV:} We applied simple thresholding in the HSV color space for sensible values for skin color \cite{kolkur2017human}.
\boldparagraph{CNN~\cite{cai2020generalizing}:} 
This refers to the bayesian CNN in Cai~\etal~\cite{cai2020generalizing} that is based on RefineNet~\cite{lin2017refinenet} and a starting point for their proposed self-supervised uncertainty-guided domain model adaption (UMA).
%
\boldparagraph{UMA~\cite{cai2020generalizing}:} We retrained the bayesian CNN using our training dataset and further ran the uncertainty-guided model adaption (UMA) on the validation dataset.
The results are recorded in \tabref{tab:results_seg}.
The performance of the CNN and the UMA is rather similar, because the training and validation dataset are from the same data distribution and the domain adaption is limited.
To make correct use of the UMA method, we further evaluated the segmentation qualitatively on captured real data, where the UMA is based on the obtained video frames.
\figref{fig:segmentation} visualizes input image crops and the respective predicted segmentation masks of each method.
Contrasting with the color thresholding, all data driven methods correctly learned to differentiate between hand and arm.
Furthermore, they are not impacted by similarly colored objects like the world map or floor.
Our method stands out at segmenting fingers, as both the CNN and UMA method fail to segment them in detail.

\subsection{Stereo extension}

For the stereo case, the training consists of 2 epochs of synthetic adaption and 8 epochs of learning the stereo weights, where we linearly decay the learning rate over the last 4 epochs.
We used Adam~\cite{kingma2014adam} with initial learning rate of $0.0002$ and $\beta_1\!=\!0.9, \beta_2\!=\!0.99$.
For the experiment in Tab.~\ref{tab:monocular_stereo} we created a 70/30 split of our synthetic stereo dataset and only used the right image for the monocular validation (the same synthetic adapted HandNet was used for comparability). 
\begin{table}[tb]
\vspace{-6pt}
\centering
\footnotesize
\setlength{\tabcolsep}{7pt}
\begin{tabular}{llll}
\toprule
Version   & MPVE (PA)$\downarrow$ & F@5mm (PA)$\uparrow$   & F@15mm (PA)$\uparrow$    \\\midrule
Monocular & 1.65          & 0.352          & 0.860          \\
Stereo    & \textbf{0.94} & \textbf{0.578} & \textbf{0.967} \\
\bottomrule
\end{tabular}
\vspace{1pt}
\caption{\textbf{Stereo case.} The stereo version greatly outperforms the monocular, showing that multiple views can resolve ambiguities. }
\label{tab:monocular_stereo}
\end{table}


\section{Discussion and Conlusion}
We argue that our hybrid approach is more expressive and personalized than work relying on the MANO hand model alone and also more efficient and robust than methods that directly try to infer a mesh without an underlying hand model.
The experiments show that our approach outperforms state-of-the-art methods in monocular RGB hand pose and shape estimation.
The integrated hand segmentation network exhibits state-of-the-art performance, which enables qualitative improvements via the proposed fine grained test time optimization.
The accurate mesh-image alignments allows for texturing, which yields visually pleasing personalized hand meshes.
Additionally, we demonstrated the usefulness of the stereo extension for increased robustness.
Due to low computational requirements, our method can be used in real-time applications.

For future work, we believe that the monocular and stereo hand mesh estimation could be further improved.
This includes the creation of new and better real datasets, as they appear to be a limiting factor especially for the challenging egocentric perspective.
In addition, the realistic appearance component could be further studied.
Due to the fine grained optimization, we were able to project the texture map, but generative approaches for the texture map creation could also be studied and lead to fruitful results.
Further, the intrinsic decomposition of texture maps for adaptive relighting could be explored.

\noindent
\begin{minipage}{\columnwidth}
	\vspace{3pt}
	\footnotesize
	\noindent
	\textbf{Acknowledgments.}~
	This research was partly supported by Innosuisse funding (Grant No.~34475.1 IP-ICT) and a research grant by FIFA.
\end{minipage}


{\small
\bibliographystyle{ieee_fullname}
\bibliography{egbib}
}


\clearpage
\onecolumn



\section*{Supplementary material}

# — Supplementary Material —
## Realistic Hands: A Hybrid Model for 3D Hand Reconstruction

Michael Seeber[1]    Roi Poranne[2]    Marc Polleyfeys[1,4]    Martin R. Oswald[1,3]

[1]ETH Zurich    [2]University of Haifa    [3]University of Amsterdam    [4]Microsoft

**Abstract.** This supplementary material provides additional details on the network architecture of our method (Sec. 1, its extension to stereo (Sec. 2, and further information about the synthetic stereo dataset (Sec. 3).

## 1. Architecture Details

**Encoder.**

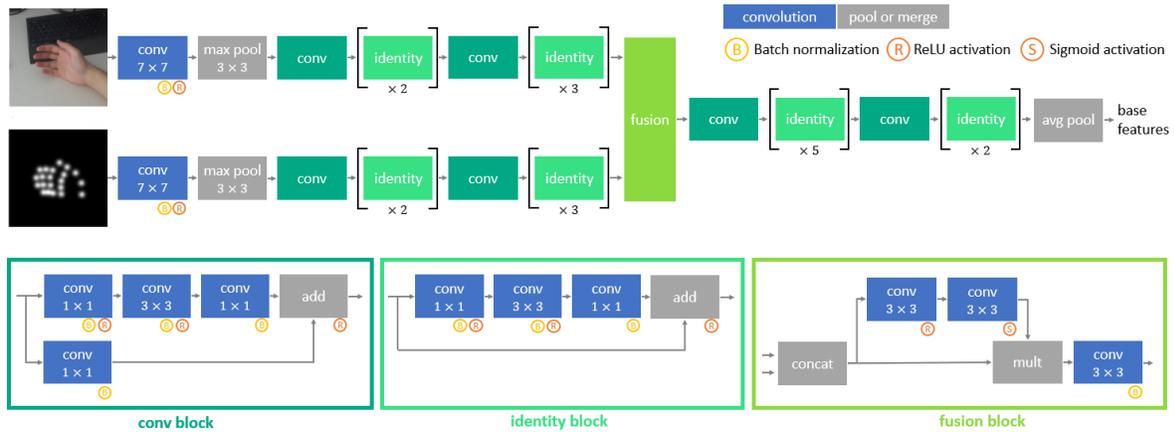

Figure 1. **Encoder Architecture.** Proposed ResNet50 based encoder with self-supervised model adaption block for modality fusion.

**Decoder.**

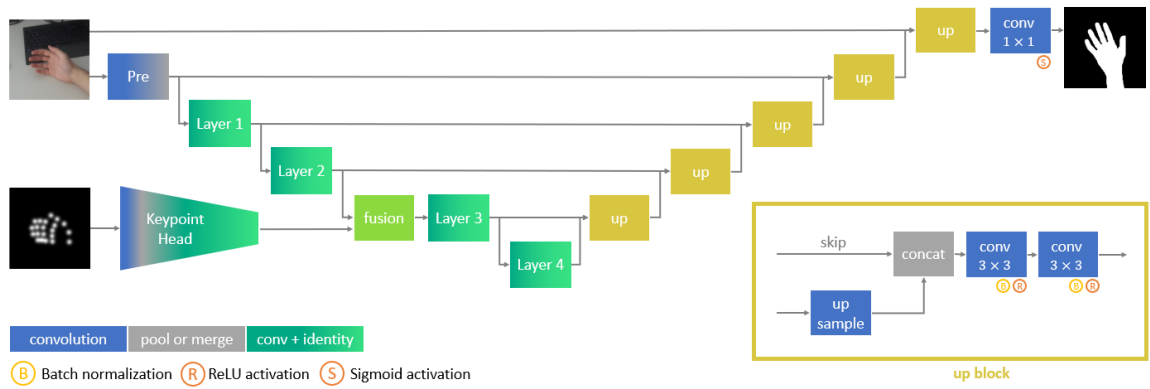

Figure 2. **Decoder Architecture.** Proposed segmentation decoder architecture inspired by U-Net with skip connections from the encoder.

**Regressor.**

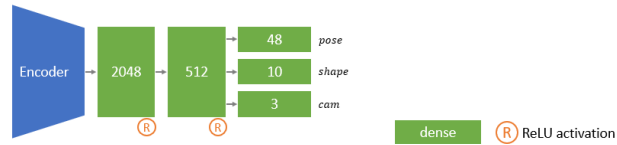

Figure 3. **Regressor Architecture.** Proposed MANO regressor component of our architecture consisting of dense layers.

**Summary.**

| Layer | | Output Size | Input Channels | Output Channels |
|---|---|---|---|---|
| ImgHead | Down1 | [112,112] | 3 | 64 |
| | Down2 | [56,56] | 64 | 256 |
| | Down3 | [28,28] | 256 | 512 |
| KpHead | Down1 | [112,112] | 3 | 64 |
| | Down2 | [56,56] | 64 | 256 |
| | Down3 | [28,28] | 256 | 512 |
| Fusion | Conv2d | [28,28] | 1024 | 128 |
| | Conv2d | [28,28] | 128 | 1024 |
| | Conv2d | [28,28] | 1024 | 512 |
| Tail | Down4 | [14,14] | 512 | 1024 |
| | Down5 | [7,7] | 1024 | 2048 |
| Segment | Up1 | [14,14] | 2048 | 1024 |
| | Up2 | [28,28] | 1024 | 512 |
| | Up3 | [56,56] | 512 | 256 |
| | Up4 | [112,112] | 256 | 128 |
| | Up5 | [224,224] | 128 | 64 |
| | Conv2d | [224,224] | 64 | 1 |
| Regress | Base | | 2048 | 512 |
| | Pose | | 512 | 48 |
| | Shape | | 512 | 10 |
| | Cam | | 512 | 3 |
| Stereo | Weights | | 1024 | 48 |

Table 1. Output size and input/output channels of our architecture.

## 2. Stereo Extension

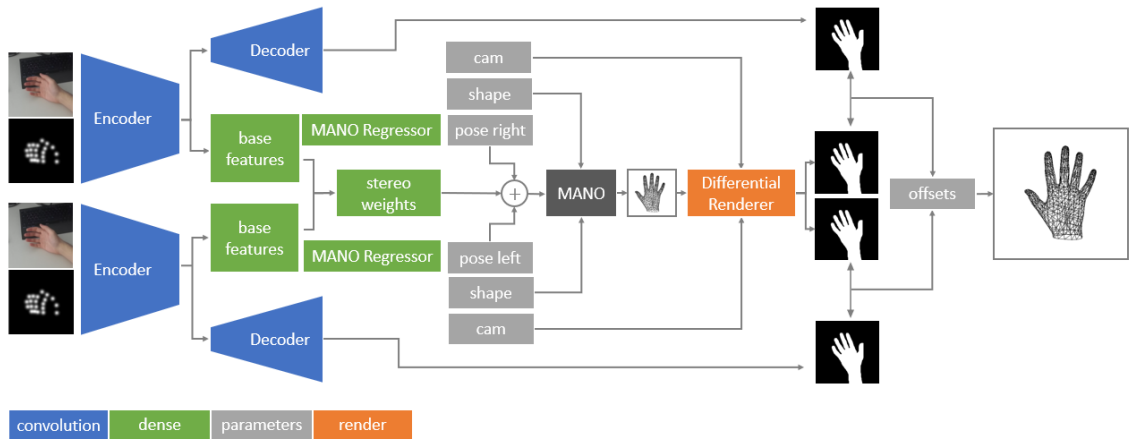

Figure 4. **Stereo Extension.** We duplicate the monocular architecture and fuse the obtained predictions through learned stereo weights. The second viewpoint is also utilized during differential rendering based test-time refinement by jointly optimizing over both views.

## 3. Synthetic Stereo Dataset

To render realistic images of embodied hands, we modified the publicly available code for the generation of the synthetic ObMan [2] to our needs. The original ObMan dataset is concerned with modeling hands grasping objects so that hand-object interactions can be studied. Our additions include the introduction of a stereo camera resembling the properties of the ZED stereo camera[1] that was used for our experiments. This includes replicating the inter-ocular distance as well as sensor and lens properties. While ObMan uses a grasp database to obtain poses, we rely on random MANO parameters to cover a more general pose space. In order to only receive likely hand articulations, we randomly sample from the PCA space with six components. The shape parameters are also uniformly selected. In order to learn the segmentation of the boundary regions between hands and arms we render the full body. This is achieved similar to [2] by converting the random hand pose to the SMPL+H [4] model which integrates MANO to the statistical body model SMPL [3]. For egocentric viewpoints, the back of the hand is more often visible than the palm of the hand. To cope with this imbalance, we rotate the global orientation of the body, such that our camera captures relevant views. Furthermore, we translate the hand root joint to align with the camera's optical center. The images are rendered using Blender [1], with the background being sampled from the living room category of the LSUN dataset [5]. We output a left and right RGB image, a left and right segmentation binary mask, as well as annotations that include hand vertices, 2D joints, 3D joints along with shape, pose and camera parameters.

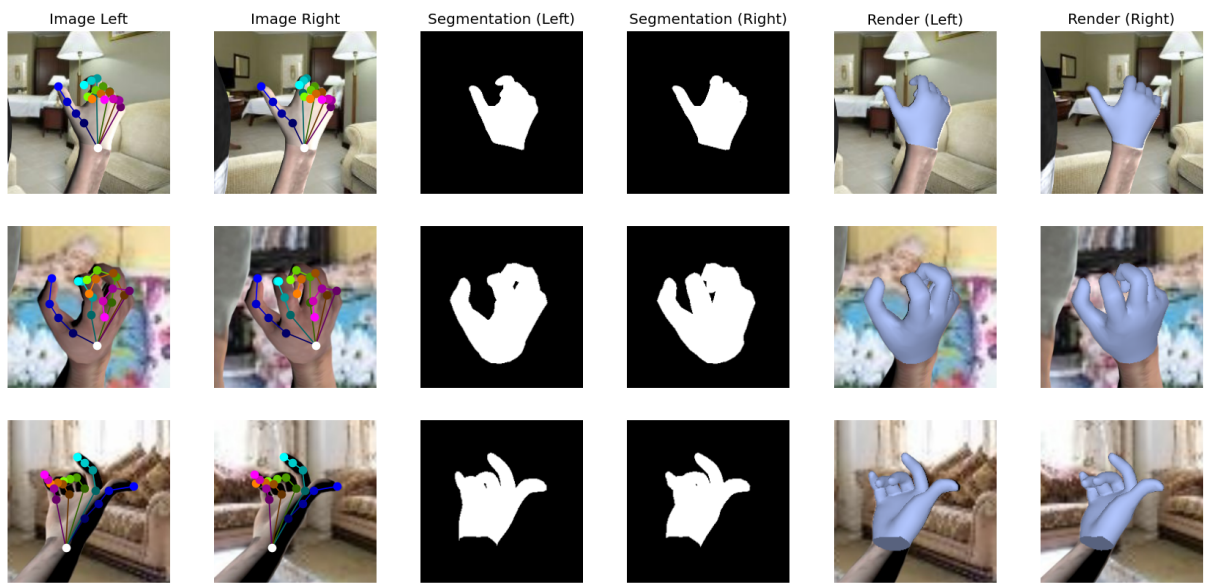

Figure 5. **Samples of our synthetic stereo dataset.** Extracted hand crops with groundtruth annotations consisting of 2D keypoints, segmentation mask and hand mesh vertices.

---

[1] https://www.stereolabs.com/zed/



\end{document}